\documentclass[11pt,a4paper]{article}
\usepackage[hyperref]{emnlp2020}
\usepackage{times}
\usepackage{latexsym}

\usepackage{adjustbox}
\usepackage{booktabs}
\usepackage{tabularx}
\usepackage{multirow}
\usepackage{siunitx}
\usepackage{makecell}
\usepackage{amsmath}
\usepackage{subcaption}
\usepackage[normalem]{ulem}
\usepackage{xspace}
\usepackage{xcolor}
\usepackage{framed}
\usepackage{soul}
\newcommand{\cc}{\textsc{cc}\xspace}
\newcommand{\reddit}{\textsc{reddit}\xspace}

\usepackage{microtype}

\aclfinalcopy 

\title{Aspect-Controlled Neural Argument Generation}

\author{Benjamin Schiller and Johannes Daxenberger and Iryna Gurevych \\
  Ubiquitous Knowledge Processing Lab (UKP-TUDA) \\
  Department of Computer Science, Technische Universit{\"a}t Darmstadt \\
  \url{www.ukp.tu-darmstadt.de} }

\date{}

\begin{document}
\maketitle
\begin{abstract}
We rely on arguments in our daily lives to deliver our opinions and base them on evidence, making them more convincing in turn. However, finding and formulating arguments can be challenging. In this work, we train a language model for argument generation that can be controlled on a fine-grained level to generate sentence-level arguments for a given topic, stance, and aspect. We define argument aspect detection as a necessary method to allow this fine-granular control and crowdsource a dataset with 5,032 arguments annotated with aspects. Our evaluation shows that our generation model is able to generate high-quality, aspect-specific arguments. Moreover, these arguments can be used to improve the performance of stance detection models via data augmentation and to generate counter-arguments. We publish all datasets and code to fine-tune the language model.\footnote{\url{https://github.com/UKPLab/controlled-argument-generation}}
\end{abstract}

\section{Introduction}
Language models \citep{bengio2003neural} allow to generate text through learned distributions of a language and have been applied to a variety of areas like machine translation \citep{Bahdanau2015}, summarization \citep{Paulus2018}, or dialogue systems \citep{Wen2017}. 
A rather new field for these models is the task of producing text with argumentative content \citep{wang-ling-2016-neural}. 
Current argument generation models, however, produce lengthy texts and allow the user little 
control over the aspect the argument should address \citep{hua-etal-2019-argument-generation,Hua2018}. 
We believe that argument generation can be enhanced by allowing for such a fine-grained control and by limiting the argument to a single but concise sentence.

Controllable language models like the CTRL \citep{Keskar2019} allow to condition the model at training time to certain control codes. At inference, these can be used to direct the model's output in regard to content or style. 
We build upon this architecture to control argument generation based solely on a given topic, stance, and argument aspect. For instance, to enforce focus on the aspect of \textit{cancer} for the topic of \textit{nuclear energy}, we input a control code ``\textit{Nuclear Energy CON cancer}'' that creates a contra argument discussing this aspect, for instance: ``\textit{Studies show that people living next to nuclear power plants have a higher risk of developing cancer.}''.
\begin{figure*}
\centering
{\includegraphics[width=0.75\textwidth]{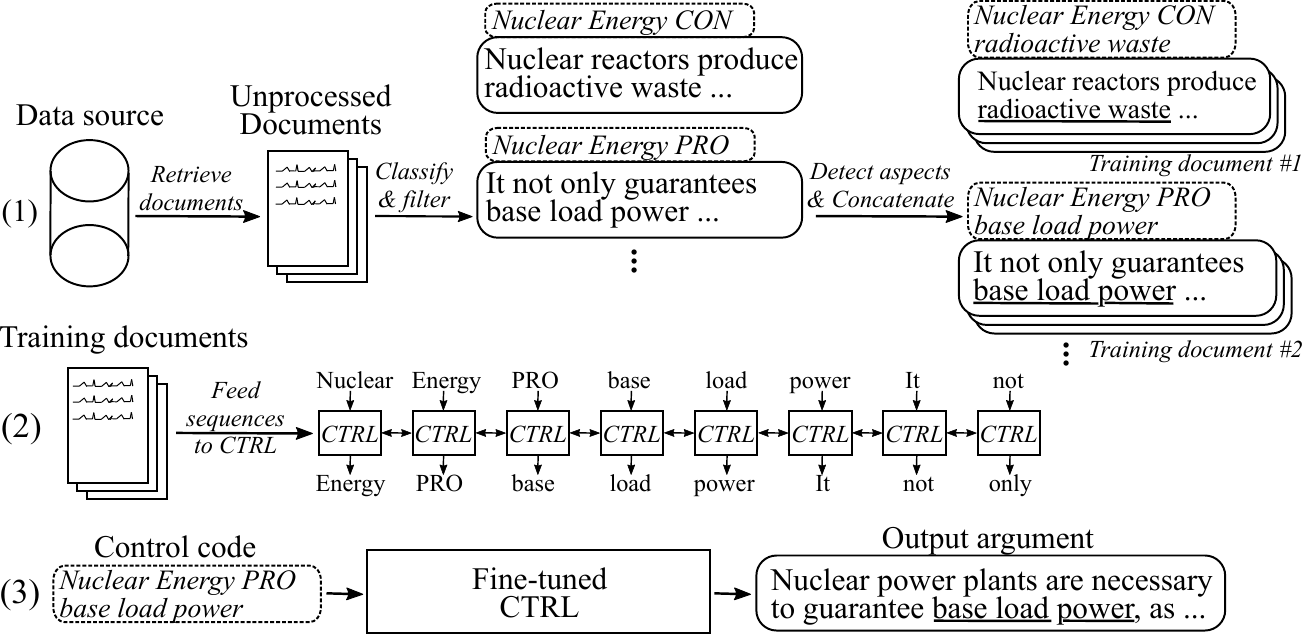}}
\caption{\label{fig:pipeline}Overview of the argument generation pipeline. (1) Gathering data from large data sources. All sentences are split, classified, and their aspects are detected. Arguments with the same topic, stance, and aspect ($\widehat{=}$ control code) are concatenated into training documents. (2) The model is fine-tuned on each training document with the control code prepended to each input sequence. (3) At inference, the model only needs a control code to generate an argument that follows the given control code.}
\end{figure*}

To obtain control codes from training data, we pre-define a set of topics to retrieve documents for and rely on an existing stance detection model to classify whether a sentence argues in favor (\textit{pro}) or against (\textit{con}) the given topic \cite{stab-etal-2018-argumentext}. 
Regarding argument aspect detection, however, past work has two drawbacks: it either uses simple rule-based extraction of verb- and noun-phrases \citep{Fujii:2006:SSV:1654641.1654644} or the definition of aspects is based on target-concepts located within the same sentence \citep{gemechu-reed-2019-decompositional}.
Aspects as we require and define them are not bound to any part-of-speech tag and (1) hold the core reason upon which the conclusion/evidence is built and (2) encode the stance towards a general but not necessarily explicitly mentioned topic the argument discusses. For instance:

\begin{center}\noindent\framebox{\small
  \begin{minipage}{0.9\linewidth}
\textbf{Topic}: \textit{Nuclear Energy} \\
\textbf{Argument}:  \textit{Nuclear reactors produce hazardous \underline{radioactive waste} and can easily be targeted by \underline{terrorist attacks}.}
  \end{minipage}%
}
\end{center}
The evidence of this argument is based upon the two underlined aspects. In addition, they encode a negative stance towards the topic and the topic is not mentioned explicitly in the sentence. 

Our final controlled argument generation pipeline (see Figure \ref{fig:pipeline}) works as follows: 
(1) We gather several million documents for eight different topics from two large data sources. All sentences are classified into pro-, con-, and non-arguments. We detect aspects of all arguments with a model trained on a novel dataset and concatenate arguments with the same topic, stance, and aspect into training documents.
(2) We use the collected classified data to condition a CTRL model on the topics, stances, and aspects of all gathered arguments. (3) At inference, passing the control code \textit{[Topic] [Stance] [Aspect]} will generate an argument that follows these commands.

Our evaluation shows that the model is able to produce aspect-specific, high-quality arguments that can be used to improve stance detection models or to create counter-arguments.
The contributions are as follows: (1) We adapt and fine-tune the CTRL model for aspect-specific neural argument generation. (2) We show that detecting argument aspects and conditioning the generation model on them are necessary steps to control the model's training process and its perspective while generating. (3) We propose novel methods to evaluate the quality of (controllable) argument generation models. (4) We develop a new scheme to annotate argument aspects and release a dataset with 5,032 samples.

\section{Related Work}
\textbf{Argument Aspect Detection}
As of today, there has not been much work on the specific field of argument aspect detection. Early work by \citet{Fujii:2006:SSV:1654641.1654644} focuses mainly on Japanese and restricts aspects to noun- and verb-phrases to extract them via hand-crafted rules. \citet{bilu-etal-2019-argument} define commonplace arguments that are valid in several situations for specified actions (e.g. ``ban'') and topics (e.g. ``smoking''). These actions are similar to aspects, but limited in number and manually defined. \citet{gemechu-reed-2019-decompositional} detect, amongst others, concepts and aspects in arguments with models trained on expert annotations. However, in their definition, aspects have to point to the target concept mentioned in the argument. In our definition, aspects refer to a general topic which is not necessarily part of the sentence and our annotation scheme is applicable by non-experts.

The concept of framing dimensions \citep{Boydstun2014TrackingTD} is close to argument aspects. In the field of argument mining, \citet{ajjour-etal-2019-modeling} have recently applied frames to label argument clusters. Yet, their method does not allow to detect frames. \citet{naderi-hirst-2017-classifying} present a method to automatically classify frames in news articles on the sentence-level. However, their approach is limited to detect fifteen frames that represent high-level concepts, whereas we operate on the token-level and identify fine-grained aspects that are explicitly mentioned in individual arguments.

\noindent\textbf{Argument Generation} 
Early approaches to argument generation rely on rules from argumentation theory and user preference models \citep{carenini2006generating,zukerman1998bayesian}. In a more recent work, \citet{sato-etal-2015-end} construct rules to find arguments in a large data source, which are then filtered and ordered with a neural network based ranker. \citet{stein:2019z} use argumentative discourse units (major claims, pro and con statements) to assemble argumentative texts with a clustering and regression approach to label, rank, and arrange the units. 
However, most of these approaches rely on hand-crafted features and thus do not generalize well. Moreover, they all require permanent access to large data sources and are not able to generate new arguments. 

Only recently, the research on generating arguments with language models gained more attention.
\citet{hua-wang-2019-sentence} use a sequence to sequence model \citep{sutskever2014sequence} that adopts the attention model of \citet{Bahdanau2015}. To generate argumentative text, the model attends to the input statement and keyphrases automatically extracted for each input from Wikipedia and news articles.
Other work focuses on generating argumentative dialogue \citep{le-etal-2018-dave} and counter-arguments \citep{hidey-mckeown-2019-fixed,hua-etal-2019-argument-generation} based on a given input sentence, or on generating summaries from a set of arguments \citep{wang-ling-2016-neural}.
In contrast, we train a language model that does not require a sentence-level input for generation and allows for direct control over the aspect of the produced argument. 

The idea that comes closest to ours is the Plug and Play Language Model by \citet{Dathathri2019}. They train two models that control the sentiment and topic of the output of a pre-trained language model at inference.
However, they do not create arguments and it is unclear whether their approach can be used to control aspects of arguments.
We show that argument generation requires the concept of argument aspects to shape the produced argument's perspective and to allow for diverse arguments for a topic of interest.

\section{Processing: Aspect Detection}\label{sec:aspect_detection}
Argument aspect detection is a necessary processing step in our argument generation pipeline, as it allows for a fine-grained control of the argument generation process. 
We create a new dataset for this task, as existing approaches either rely on coarse-grained frames or cannot be applied by non-expert annonators in a scalable manner.

\subsection{Dataset Creation}\label{sec:data_creation_analysis}
We base our new aspect detection dataset on the UKP Sentential Argument Mining Corpus (UKP-Corpus) by \citet{Stab2018b}, as it already contains sentence-level arguments and the other two control codes we aim to use: topics and stance-labels. More precisely, it contains 25,474 manually labelled arguments for eight controversial topics in English. 
Each sample consists of a topic and a sentence, labelled as either being supporting, attacking, or no argument towards the given topic. As we are only interested in arguments, we do not consider the non-argumentative sentences.

\noindent \textbf{Step 1: Preliminary annotations} To ensure the feasibility of creating a dataset for this difficult task, two experts (a post-doctoral researcher and an undergraduate student with NLP background) independently annotate 800 random samples (from four topics, 200 per topic) taken from the UKP-Corpus. 
The annotations are binary and on token-level, where multiple spans of tokens could be selected as aspects. 
The resulting inter-annotator agreement of this study is Krippendorff's $\alpha_u = .38$. While this shows that the task is generally feasible, the agreement on exact token spans is rather low. Hence, in the following steps, we reduce the complexity of the annotation task.

\noindent \textbf{Step 2: Annotation scheme} Instead of free span-level annotations, we present annotators with a ranked list of aspect recommendations. To generate meaningful recommendations, we train a ranking model using the preliminary annotations (Step~1). 

\noindent \textbf{Step 2a: Data preparation for ranking} To create training data for the ranker, we use a simple heuristic to calculate scores between 0 and 1 for all N-grams of a sentence by dividing the number of aspect tokens within an N-gram by its length $N$: $\dfrac{\#\,\,aspect\,\,tokens}{N} \in [0,1]$.
Our analysis reveals that 96\% (783 of 814) of all aspects in the preliminary annotation dataset only contain one to four tokens. 
We thus decide to ignore all candidates with more than four tokens.
No other limitations or filtering mechanisms are applied.

\begin{table*}
\centering 
\small{
\begin{tabular}{ll}
\Xhline{2\arrayrulewidth}

\textbf{Topic} & \textbf{Five most frequent aspects (frequency)} \\\hline
Gun control & \makecell[l]{right (30), protect (18), background checks (17), gun violence (14), criminal (13)}\\

Death penalty & \makecell[l]{cost (16), innocent (12), retribution (10), murder rate (9), deterrent (8)}\\

Abortion & \makecell[l]{right (21), pain (10), choice (10), right to life (9), risk (9)}\\

Marijuana legalization & \makecell[l]{dangerous (16), cost (13), risk (12), harm (10), black market (9)}\\

Nuclear energy & \makecell[l]{cost (32), accident (24), waste (22), risk (16), dangerous (13)}\\

School uniforms & \makecell[l]{bullying (14), individuality (12), cost (11), safety (11), discipline (9)}\\

Minimum wage & \makecell[l]{poverty (28), cost (26), economy (17), wage (14), unemployment (13)}\\

Cloning & \makecell[l]{human dignity (17), disease (12), individuality (11), unethical (11), stem cell (10)}\\
\hline

General aspects & \makecell[l]{dangerous (in 8 of 8 topics), cost / life / risk / safety (in 7 of 8 topics)}\\

\Xhline{2\arrayrulewidth}
\end{tabular}   
}
\caption{The five most frequent aspects for each topic.}
\label{tbl:most_frequent_aspects}
\end{table*}

\begin{table}
\centering 
\resizebox{0.45\textwidth}{!}{
\def\arraystretch{1.3}
\begin{tabular}{lcccc}
\Xhline{2\arrayrulewidth}
 \textbf{Setting} & \textbf{Rec@5} & \textbf{Rec@10} & \textbf{Rec@15} & \textbf{Rec@20} \\\hline
\textbf{In-topic} & 0.7701 & 0.8468 & 0.8661 & 0.8925 \\
\textbf{Cross-topic} & 0.5951 & 0.7415 & 0.8164 & 0.8630 \\
\Xhline{2\arrayrulewidth}
\end{tabular}
}
\caption{In- and cross-topic Recall@k of the ranker used for aspect candidate recommendations.}
\label{tbl:recall_pilot}
\end{table}

\noindent \textbf{Step 2b: Training the ranker} We use BERT \citep{DBLP:journals/corr/abs-1810-04805} and MT-DNN \citep{liu2019mt-dnn} (base and large) to train a ranker. For training, we create five splits: (1) one in-topic split trained on a random subset from all four topics and (2) four cross-topic splits that are trained via leave-one-topic-out strategy.
The cross-topic setup allows us to estimate the ranker's performance on unseen topics of the UKP-Corpus.

A single data sample is represented by an argument and an 1- to 4-gram of this argument, separated by the BERT architecture's [SEP] token. 
This technique expands the 800 original samples of the dataset to around 80,336. 
We use the mean squared error as loss and take the recall@k to compare the models. 
The in- and cross-topic results of the best-performing model (MT-DNN$_{BASE}$) are reported in Table \ref{tbl:recall_pilot}. All results are the average over runs with five different seeds (and over all four splits for the cross-topic experiments).

\noindent \textbf{Step 2c: Creating the annotation data} 
For the four topics that are part of the preliminary annotation dataset, we use the in-topic model to predict aspects of 629 random unseen arguments from the UKP-Corpus. For the other four topics of the UKP-Corpus, we choose the best cross-topic model to predict aspects for the same amount of samples. 
To keep a recall of at least 80\%, we choose the ten and fifteen highest-ranked aspect candidates for samples as predicted by the in-topic and cross-topic model, respectively. 
We remove aspect candidates that include punctuation, begin or end with stopwords, or contain digits.

\noindent \textbf{Step 3: Annotation study} We use Amazon Mechanical Turk to annotate each sample by eight different workers. Based on a subset of 232 samples, we compute an $\alpha_u$ of .67 between crowdworkers and experts (three doctoral researchers). Compared to the initial study, the new approach increases the inter-annotator agreement between experts by approx. 12 points.
Based on this promising result, we create a dataset of 5,032 high-quality samples that are labelled with aspects and the stance labels from the original dataset.
A more detailed explanation of the process, as well as the guidelines and screenshots of the annotation study design can be found in \ref{sec:annotation_study}. Exemplary, we show the most frequent (lemmatized) aspects that appear in all topics in Table \ref{tbl:most_frequent_aspects}.

\begin{table}
\centering 
\resizebox{0.44\textwidth}{!}{
\def\arraystretch{1.3}
\begin{tabular}{lccc}
\Xhline{2\arrayrulewidth}
 \textbf{Model} & \textbf{F$_1$ macro} & \textbf{Precision} & \textbf{Recall} \\\hline
 \textbf{Majority (baseline)} & .3085 & .2871 & .3333 \\ 
\textbf{Ranker (baseline)} & .6522 & .6685 & .6474 \\ 
\textbf{BERT$_{BASE}$} & .6980 & .6927 & \textbf{.7040} \\
\textbf{BERT$_{LARGE}$} & \textbf{.7100} & \textbf{.7240} & .6993 \\\Xhline{2\arrayrulewidth}
\end{tabular}
}
\caption{Test set results of the models for aspect detection. Majority only predicts class O.}
\label{tbl:baselines_aspect}
\end{table}

\subsection{Evaluation}\label{sec:aspect_eval}
We create a cross-topic split with the data of two topics as test set (\textit{gun control}, \textit{school uniforms}), one topic as dev set (\textit{death penalty}), and the remaining topics as train set and evaluate two models with it. First, we use the ranking approach described in Step 2a-2b to fine-tune MT-DNN$_{BASE}$ on the newly generated data. At inference, we choose the top $T$ aspects for each argument as candidates. We tune $T$ on the dev set and find $T=2$ to be the best choice. 
Second, we use BERT for sequence tagging \citep{Wolf2019HuggingFacesTS} and label all tokens of the samples with BIO tags. 
As previously done with the ranker, we test this model with BERT and MT-DNN weights and find BERT$_{LARGE}$ to be the best choice. We flatten the predictions for all test samples and calculate the F$_1$, Precision, and Recall macro scores. 
Note, that this is a very strict method, as it does not consider overlapping aspects as correct prediction. 
All models are trained over five seeds and the results are reported in Table \ref{tbl:baselines_aspect}.

BERT$_{LARGE}$ predicts classes B and I  with an F$_1$ of .65 and .53, hence aspects with more than one token are less well identified. 
A difference is to be expected, as the class balance of B's to I's is 2768 to 2103. 
While the ranker performs worse based on the shown metrics, it has a slightly higher recall for class I. 
We assume this is due to the fact that it generally ranks aspects with more than one token on top, i.e. there will often be at least one or more I's in the prediction. 
In contrast to that, BERT$_{LARGE}$ focuses more on shorter aspects, which is also in accordance with the average aspect length of 1.8 tokens per aspect in the dataset. In total, BERT$_{LARGE}$ outperforms the simpler Ranker baseline by almost 6 percentage points in F$_1$ macro.

\section{Data Retrieval Pipeline for Argument Generation}\label{sec:pre_arggen}
In this section, we describe the data retrieval and preprocessing for the argument generation pipeline. We aim to train a model that is able to transfer argumentative information concisely within a single sentence. We lean onto \citet{Stab2018b} who define such an argument as the combination of a topic and a sentence holding evidence with a specific stance towards this topic. Consequently, the following preprocessing steps ultimately target retrieval and classification of sentences.
To evaluate different data sources, we use a dump from Common-Crawl\footnote{\url{https://commoncrawl.org}} (\cc) and Reddit comments\footnote{\url{https://files.pushshift.io/reddit/comments/}} (\reddit) to train two separate generation models. The \cc dump is from 2016 and contains (after preprocessing and deduplication) 331M documents (3.6TB). The \reddit dump contains 2.5B documents (1.6TB) from December 2012 to May 2019.

\noindent\textbf{Document Retrieval} We index \reddit and \cc with ElasticSearch\footnote{\url{https://www.elastic.co}} and, for both, gather up to 1.5M documents for each of the eight topics of the UKP-Corpus. To increase the search results, we add synonyms (see \ref{sec:synonyms}) for most topics. 

\noindent\textbf{Argument and Stance Classification} We split the sentences of all documents and remove duplicates. We notice that many sentences are not relevant in regard to the document's topic. To enforce topic-relevance, we decide to filter out all sentences that do not contain at least one token of the respective topics or defined synonyms (see \ref{sec:synonyms}).
We use the ArgumenText API's\footnote{\url{https://api.argumentsearch.com}} argument and stance classification models \cite{stab-etal-2018-argumentext} to classify all sentences into \textit{arguments} or \textit{non-arguments} (F$_1$ macro $=.7384$), and all remaining arguments into \textit{pro} or \textit{con} in regard to the topic (F$_1$ macro $=.7661$).

\noindent\textbf{Aspect Detection}
We detect aspects on all remaining arguments. To speed up the detection on millions of sentences, we use BERT$_{BASE}$ instead of BERT$_{LARGE}$ (see Table \ref{tbl:baselines_aspect}). 

\noindent\textbf{Training Document Generation}\label{sec:document_gen} To generate training documents for the model, we concatenate all arguments that have the same topic, stance, and aspect (i.e. the same control code). In addition, we aggregate all arguments that include an aspect with the same stem into the same document (e.g. arguments with \textit{cost} and \textit{costs} as aspect). 
To cope with limited hardware resources, we restrict the total number of arguments for each topic and stance to 100,000 (i.e. 1.6M over all eight topics).
Moreover, as some aspects dominate by means of quantity of related arguments and others appear only rarely, we set an upper and lower bound of 1,500 and 15 arguments to each document. 

\section{Model, Training, and Analysis}\label{sec:gen_training}
In the following, we describe the architecture and the training process of the generation model and analyze it in comparison to a retrieval-based model.

\noindent\textbf{Model}
The goal of a statistical language model is to learn the conditional probability of the next word given all (or a subset of) the previous ones \citep{bengio2003neural}. That is, for a sequence of tokens $x = (x_1, ..., x_n)$, the model learns $p(x_i|x_{<i})$ where $x_i$ is the $i$-th word of sequence $x$. 
For this work, we use the Conditional Transformer Language Model (CTRL) by \citet{Keskar2019}. It has shown to produce high quality text and it can be adapted for conditioning on the control codes we aim to use, without the need of pre-training the weights from scratch.
Formally, the CTRL adds an extra condition to each sequence by prepending a control code $c$, hence learning $p(x_i|x_{<i},c)$. The control code is represented by a single token and can then be used to direct the model output at inference. 
Architecture-wise, the CTRL is built on a transformer-based sequence to sequence architecture \citep{vaswani2017attention}. The model was trained on 140GB of data from several large resources like Wikipedia, subreddits, and news data. We extend the model from its previous limit of a single-token control code to accept multiple tokens. We use the pre-trained weights with a sequence length of 256 and fine-tune them on our own data.

\noindent\textbf{Training}
We train the model on a Tesla V100 with 32 GB of Memory. We keep the default hyperparameters and only reduce the batch size to 4. All training documents are sampled randomly for training. The respective control code is prepended to each sequence of 256 subwords of a document.
The model takes around five days to train on the 1.6M training sentences.

\noindent\textbf{Generation}
At inference, we observe that for the first generated argument, the model mostly outputs very short phrases, as it tries to incorporate the control code into a meaningful start of an argument. We prevent this by adding punctuation marks after each control code (e.g. a period or colon), signaling the model to start a new sentence. In this fashion, we generate \textit{pro} and \textit{con} arguments up to the pre-defined training split size for each topic of the UKP-Corpus, resulting in 7,991 newly generated argument samples. We do this for the model based on the \reddit and \cc data and use the generated arguments as a basis for the following analysis and evaluation methods. Examples of generated arguments for both models can be found in tables \ref{tbl_ex_quality_short} and \ref{tbl_ex_counter_arg_short} as part of the evaluation (see Section \ref{sec:eval}).

\noindent\textbf{Results and Analysis}
As an upper bound, we compare the generation models to a retrieval approach, which returns all arguments stored for a given topic, stance, and aspect from the gathered training documents (see Section \ref{sec:pre_arggen}).
Both the retrieval and generation approaches are evaluated against reference data from debate portals and compared via METEOR \citep{lavie-agarwal-2007-meteor} and ROUGE-L \citep{lin-2004-rouge} metrics. 
The retrieval approach has the advantage, as the arguments are of human origin and aspects are always explicitly stated within a belonging argument.

The reference data was crawled from two debate portals\footnote{procon.org and idebate.org} and consists of pro- and con-paragraphs discussing the eight topics of the UKP-Corpus. As the paragraphs may also include non-arguments, we filter these out by classifying all sentences with the ArgumenText API into arguments and non-arguments. This leaves us with 349 pro- and 355 con-arguments over all eight topics (see \ref{sec:ref_data_stats} for more details). 
Next, we detect the aspects in these arguments. All arguments with the same aspect are then used as reference for arguments with the same aspect from the (a) generated arguments and (b) retrieval approach arguments.
The results show that METEOR and ROUGE-L are only approx. 0.5-1.1 and 2.7-2.9 points lower for the generation models, respectively (see Table \ref{tbl_human_eval}). It not only shows the strength of the architecture, but also the success in generating sound aspect-specific arguments with our approach.

\begin{table}
\centering 
\resizebox{0.4\textwidth}{!}{
\def\arraystretch{1.3}
\begin{tabular}{lcc}
\Xhline{2\arrayrulewidth}
\textbf{Model}  & \textbf{METEOR} &  \textbf{ROUGE-L} \\\hline
       Retrieval (\cc)  & \textbf{17.85} & \textbf{14.72} \\
       
       CTRL (\cc) & 16.80 & 11.95 \\\hline
       
       Retrieval (\reddit)  & \textbf{17.29} & \textbf{15.26}\\
       
       CTRL (\reddit) & 16.82 & 12.34\\\Xhline{2\arrayrulewidth}
\end{tabular}
}
\caption{Comparison of retrieval and generation approach with reference data from debate portals.}\label{tbl_human_eval}
\end{table}

\section{Generation in Absence of Aspects}
To show the necessity of having prior knowledge of aspects to aggregate data for argument generation,
we create training data \textit{without} prior knowledge of aspects, train a new generation model on it, and compare it to our previous models \textit{with} prior knowledge of aspects.
Equally to the original \cc model's procedure, we gather 100,000 sentences for each stance of a topic from the \cc data. As we assume to have no knowledge about the aspects of the arguments in this scenario, we randomly sample arguments from the \cc source documents. We create training documents with numbers of arguments varying between 15 and 1500 to mimic the data generation process of the original models and train a new generation model on them. After training, we generate arguments for the new model in the same manner as previously done for the other models by using aspects contained in the training data as control codes. While the model was not conditioned on these control codes, they nevertheless appear in the training data in at least one argument.

We compare all models by verifying whether or not the aspect used for generation (including synonyms and their stems and lemmas) can be found in the generated arguments. 
For the original models conditioned on aspects, this is true in 79\% of the cases for \reddit and in 74\% of the cases for \cc. For the non-aspect model, however, it is only true in 8\% of the cases. It clearly shows the necessity to condition the model on aspects explicitly, implying the need for argument aspect detection, as the model is unable to learn generating aspect-related arguments otherwise. 
Moreover, without prior detection of aspects, we have no means for proper aggregation over aspects.
We notice that for the model without prior knowledge about aspects, 79\% of all aspects in the training data appear in only one argument. 
For these aspects, the model likely will not pick up a strong enough signal to learn them.

\begin{table*}
\small
\begin{tabularx}{\linewidth}{|X|}
\hline
\textbf{cloning CON unrespectable .} Cloning humans for reproductive purposes is unethical and unacceptable , but creating cloned embryos solely for research - which involves destroying them anyway - is downright criminal . (0.97)\\
\textbf{cloning CON disfavored .} , cliques ) to them . (0.36)\\\hline

\textbf{nuclear energy PRO safe .} In addition , we must continue developing safer technologies like small modular reactors which will help us meet our nation 's need for reliable , emission-free sources of low-emission energy [...] . (0.96)\\
\textbf{nuclear energy CON leak .} `` We are concerned about the possibility of further releases of radioactivity due to possible melting or cracking of fuel rods at the No . (0.47)\\\hline

\textbf{marijuana legalization PRO safer :} Legalizing cannabis will help reduce crime rates ( especially violent crimes ) and make society safer overall . (0.96)\\
\textbf{marijuana legalization PRO benefits .} Decrease amount of police officers needed 6 . (0.37)\\\hline

\textbf{minimum wage PRO poor :} Raising the minimum wage will not only benefit those working full time but also reduce government expenditures on social services such as food stamps [...] which disproportionately affect the poor . (0.97)\\
\textbf{minimum wage CON cost :} If you raise the price of a Big Mac to \$ 10 and then pay an extra dime or two per burger so that it 's still only \$ 9 ... well , maybe your business is n't worth saving at all [...] . (0.44)\\\hline
                                  
\end{tabularx}
\caption{Generated arguments of highest/lowest quality with \cc as data source. Bold text shows the used control code. Quality score in brackets as predicted by the argument quality model. ``[...]'' signals shortened text.}\label{tbl_ex_quality_short}
\end{table*}

\section{Evaluation}\label{sec:eval}
In this section, we evaluate the correctness and quality of the generation models (intrinsic evaluation) and their performance on exemplary tasks (extrinsic evaluation). As a basis, we use the 7,991 arguments generated in Section \ref{sec:gen_training}.
\subsection{Intrinsic Evaluation}
\subsubsection{Argument Quality} 
We introduce a novel method to evaluate generated arguments based on the argument quality detection approach proposed by \citep{gretz2019large}. They create an argument quality dataset that contains around 30,000 arguments over 71 topics.
For each argument, annotators were asked whether or not they would recommend a friend to use the displayed argument in a speech. The quality scores for each argument result from a weighted average (WA) or MACE Probability function (MACE-P) of all annotators and range between 0 (lowest quality) and 1.0 (highest quality).
We use the \textit{WA}-score as label, the same model (BERT$_{BASE}$) and hyperparameters as given in the original paper, and reproduce the reported correlations of .52 (Pearson) and .48 (Spearman) on the test dataset (averaged over five different seeds). 
The model predicts an average argument quality of .71 for the generated \reddit arguments, .75 for the training arguments of the UKP-Corpus, and even .76 for the \cc data. 
It shows that our model is able to produce arguments that are generally on a similar quality level as human arguments. For each topic, we also retrieve the generated arguments with the highest and the lowest argument quality and show a subset in Table \ref{tbl_ex_quality_short} (see \ref{sec:ex_generated_quality} for the full set).

\begin{table}
\centering 
\resizebox{0.3\textwidth}{!}{
\def\arraystretch{1.3}
\begin{tabular}{lcc}
\Xhline{2\arrayrulewidth}
\textbf{} & \textbf{\reddit} & \textbf{\cc}  \\\hline
F$_1$ \textit{pro} & 0.7828 & 0.7972 \\
F$_1$ \textit{con}  & 0.8140 & 0.8308  \\ 
\textit{none} (in \%)  & 13.8\% & 15.6\%  \\ 
\Xhline{2\arrayrulewidth}
\end{tabular}
}
\caption{Class correctness of generated arguments. }
\label{tbl_clf_gen}
\end{table}

\subsubsection{Argument and Stance Correctness}
We use the classifier of the ArgumenText API as an oracle to produce ``gold'' labels and classify all generated arguments into \textit{pro}-, \textit{con}-, and non-Arguments (\textit{none}). Labels for the generated arguments are represented by the control codes that were used to generate them. 
Table \ref{tbl_clf_gen} shows the results of the evaluation. We observe generally high F$_1$ for \textit{pro} and \textit{con} classes for both data sources. However, 13.8\% of the generated sentences based on the \reddit model and 15.6\% based on the \cc model are not classified to be arguments. 

\subsection{Extrinsic Evaluation}
\subsubsection{Enhance Stance Prediction Model}
We show that the generated arguments can be used to enhance a stance prediction model. For this, we compare three setups: (1) Training only on arguments of the UKP-Corpus, (2) training only on generated arguments, and (3) combining generated arguments and training data of the UKP-Corpus. In addition, we compare all three setups for in- and cross-topic experiments. For in-topic experiments, we train a model for each topic separately (based on the pre-defined split in the UKP-Corpus) and average the F$_1$ macros for all topics. For cross-topic experiments, we use the same setup as in Section \ref{sec:aspect_eval}: We take \textit{all samples} (i.e. train, dev, and test split) of five topics for training, all samples of one topic for development, and all samples of two topics for testing. As we previously only generated samples up to the size of the train split of a topic, we generate additional samples for the cross-topic evaluation to fill up the dev and test split portions.
For all setups, we train BERT$_{BASE}$ on 5 epochs, a batch size of 16, a learning rate of 5e-5, and average over 5 different seed runs. 

The results in Table \ref{tbl_std} show that in-topic, the generated arguments can slightly increase the performance if combined with the original data. Yet, using only generated data leads to a much lower performance, which is not unexpected, as the original data is drawn from the same source and very similar in style. However, for cross-topic experiments, using only generated arguments improves over using the original data. While we attribute this to a higher similarity to the test topics, it suggests that the generated data can be successfully used to enhance the generalization ability of BERT in cross-topic experiments. The combination of original and generated data shows barely any difference, suggesting that it does not add new information.

\begin{table}
\centering 
\resizebox{0.37\textwidth}{!}{
\def\arraystretch{1.3}
\begin{tabular}{lcc}
\Xhline{2\arrayrulewidth}
\textbf{Method}  & \textbf{\reddit} & \textbf{\cc} \\
    &  \multicolumn{2}{c}{\textbf{F$_1$ macro}}  \\\hline
 \textbf{In-topic}           &                 \\
       UKP-Corpus               & \multicolumn{2}{c}{.7037}     \\
       Only generated & .6414 & .6409 \\
       UKP-Corpus \& generated            & \textbf{.7156} & \textbf{.7172} \\
       \textbf{Cross-topic}        &                 \\
       UKP-Corpus              &   \multicolumn{2}{c}{.5948}    \\
       Only generated & .6154 & \textbf{.6130}\\
       UKP-Corpus \& generated        & \textbf{.6184} & .6120 \\\Xhline{2\arrayrulewidth}
\end{tabular}
}
\caption{Stance prediction results with BERT$_{BASE}$ on combinations of original and generated data.}\label{tbl_std}
\end{table}

\subsubsection{Counter-Argument Generation}
Drafting counter-arguments is an important skill for debating, to provide constructive feedback, and to foster critical thinking overall. We lean onto the work of \citet{wachsmuth-etal-2018-retrieval} who describe counter-arguments as discussing the same aspect as the initial argument, but with a switched stance. Hence, given our defined control codes, our model is especially fit for counter-argument generation. Unlike current models for this task, we do not require a specific dataset with argument and counter-argument pairs \citep{hidey-mckeown-2019-fixed,hua-etal-2019-argument-generation}. Also, in contrast to the model by \citet{hua-wang-2019-sentence} that implicitly integrates input-related ``Keyphrases'' into the process of counter-argument generation, our model is able to concentrate on every aspect of the input explicitly and with a separate argument, allowing for more transparency and interpretability over the process of counter-argument generation.
We show exemplary how the combination of aspect detection and controlled argument generation can be successfully leveraged to tackle this task. We manually compose an initial argument for the topics \textit{nuclear energy} and \textit{school uniforms}. Then, we automatically detect their aspects with our model and generate a counter-argument for each aspect by passing the topic, opposite stance of the original argument, and one of the aspects into the argument generation model. For both topics, the \cc generation model produces meaningful counter-arguments based on the detected aspects (see Table \ref{tbl_ex_counter_arg_short}).

\begin{table}
\small
\begin{tabularx}{\linewidth}{|X|}
\hline
\textbf{Human:} Nuclear energy produces \underline{waste} that stays \underline{radioactive} for thousands of years and pollutes the \underline{environment}.\\
\textbf{CTRL:} (nuclear energy PRO waste .) The only thing that 's going to be left over is the spent fuel
rods , which are not very radioactive and can be reprocessed [...] 
\\
\textbf{CTRL:} (nuclear energy PRO radioactive .) The radioactivity of the waste is not a problem , because it will decay away in about 300 years or so [...] 
\\
\textbf{CTRL:} (nuclear energy PRO environment .) The fact that the radioactive waste is not only stored in a safe manner , but also protected from any potential threat to human health or the environment for thousands of years [...] 
\\\hline

\textbf{Human:} School uniforms are \underline{expensive} and affect the pupil's \underline{individuality}.\\
\textbf{CTRL:} (school uniforms PRO expensive .) The cost of the uniform is not that much more than a regular shirt and shorts [...] \\ 
\textbf{CTRL:} (school uniforms PRO individuality .) The uniform is a symbol of unity and identity , which helps to foster an environment where students can feel comfortable about expressing their own individual style without being judged by others .\\\hline

\end{tabularx}
\caption{Generated counter-arguments with \cc model. Aspects in the initial argument are underlined and used for the counter-argument generation. Control code in brackets and ``[...]'' signals shortened text.}\label{tbl_ex_counter_arg_short}
\end{table}

\section{Conclusion}
We apply the concept of controlled neural text generation to the domain of argument generation. Our fine-tuned generation model is conditioned on topics, stances, and aspects and can reliably create arguments using these three control codes. We show that arguments generated with this approach are of high quality in general and can be used to improve the performance of stance detection models. Moreover, we show that our approach can successfully generate counter-arguments in a transparent and interpretable way. We defined the method of argument aspect detection for controlled argument generation and introduced a novel annotation scheme to crowdsource argument aspect annotations, resulting in a high-quality dataset ($\alpha_u$=.67). To foster research on (controlled) argument generation, we publish the model weights, as well as the data and all code necessary to fine-tune the generation model.

\section{Acknowledgements}
We thank Tilman Beck for his helpful comments and for providing the reference data for Section 5, \textit{Results and Analysis}.
This work has been supported by the German Research Foundation within the project ``Open Argument Mining'' (GU 798/25-1), associated with the Priority Program ``Robust Argumentation Machines (RATIO)'' (SPP-1999), and by the German Federal Ministry of Education and Research (BMBF) under the promotional reference 03VP02540 (ArgumenText). 

\bibliography{emnlp2020}
\bibliographystyle{acl_natbib}

\appendix

\section{Appendices}
\label{sec:appendix}
\subsection{Ethical Note}
The dangers of misuse of language models like the CTRL have been extensively discussed by its authors \cite{Keskar2019}. Our adaption of the model may be used to automatically generate arguments and counter-arguments which, in many cases, cannot be distinguished from human-made arguments. While our intentions are to support society, to foster diversity in debates, and to encourage research on this important topic, we are aware of the harmful applications this model can be used for (e.g. biasing debates into a certain direction or spreading disinformation). However, with good intentions in mind, controllable argument generation can also be used for debiasing and to make discussions more diverse. Moreover, we believe that providing access to these models is of major importance, as their development cannot be regulated and, if they are open-sourced, it also encourages the work on counter-measures to detect them \cite{varshney2020limits}.

\subsection{Aspect Detection Annotation Study Details}\label{sec:annotation_study}
For the final crowdsourcing study, we use Amazon Mechanical Turk. 
Workers had to take a qualification test, have an acceptance rate of at least 95\%, and location within the US. We paid \$7.6 per hour (minimum wage is \$7.25 per hour).
Each data sample is annotated by eight crowdworkers. In case the ranker cut off the real aspect(s) from the list of candidates, crowdworkers could select any sequence up to four tokens from a second list. The guidelines and screenshots of the annotation interface are described in \ref{sec:study_interface}.
To test the quality of the resulting annotations, we let three experts (doctoral researchers) and eight crowdworkers annotate 29 samples per topic (232 in total). The $\alpha_u$ between the experts is .493, i.e. approx. 12 points higher than in the initial study. We then use a variant of the Bayesian classifier combination IBCC \citep{DBLP:journals/corr/abs-1811-00780,kim2012bayesian} to aggregate the labels for both groups and compute an inter-annotator agreement of $\alpha_u$ = .67 between the two.
The final dataset consists of 5,032 high-quality samples that are labelled with aspects, as well as with the stance labels from the original dataset. For reproducibility of results, we create fixed splits for in- and cross-topic experiments.

\subsection{Aspect Detection Guidelines and Interface}\label{sec:study_interface}
\begin{figure*}
    \includegraphics[scale=.4]{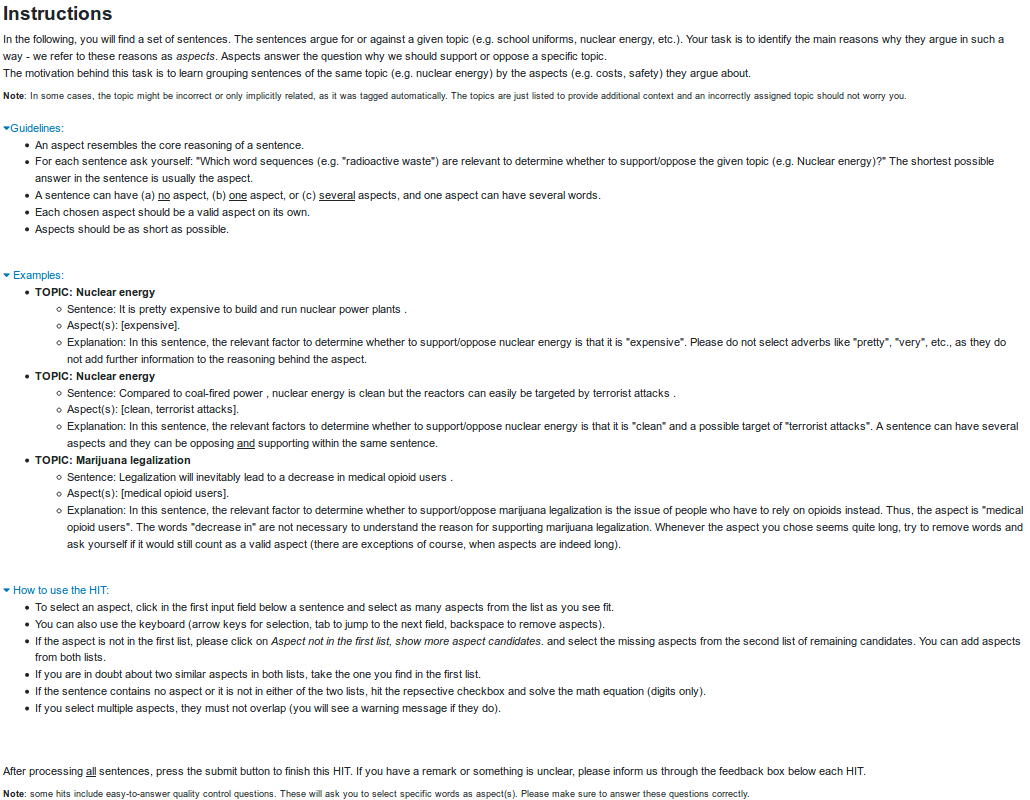}
    \caption{Guidelines for the final annotation study.}
    \label{fig:aspect_amt_guidlines}
\end{figure*}

\begin{figure*}
    \includegraphics[scale=.4]{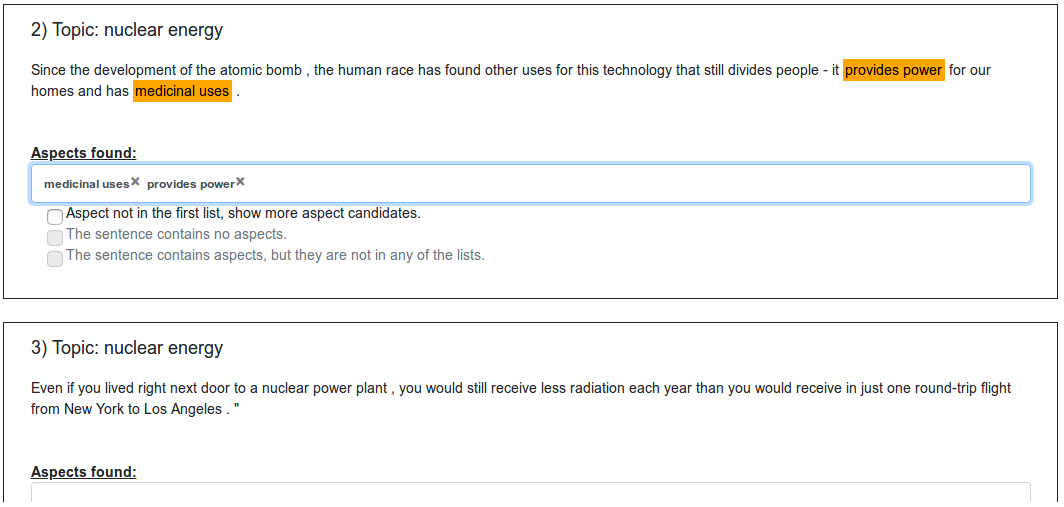}
    \caption{Example sentence of a HIT with two aspects selected.}
    \label{fig:aspect_amt_ex1}
\end{figure*}

\begin{figure*}
    \includegraphics[scale=.4]{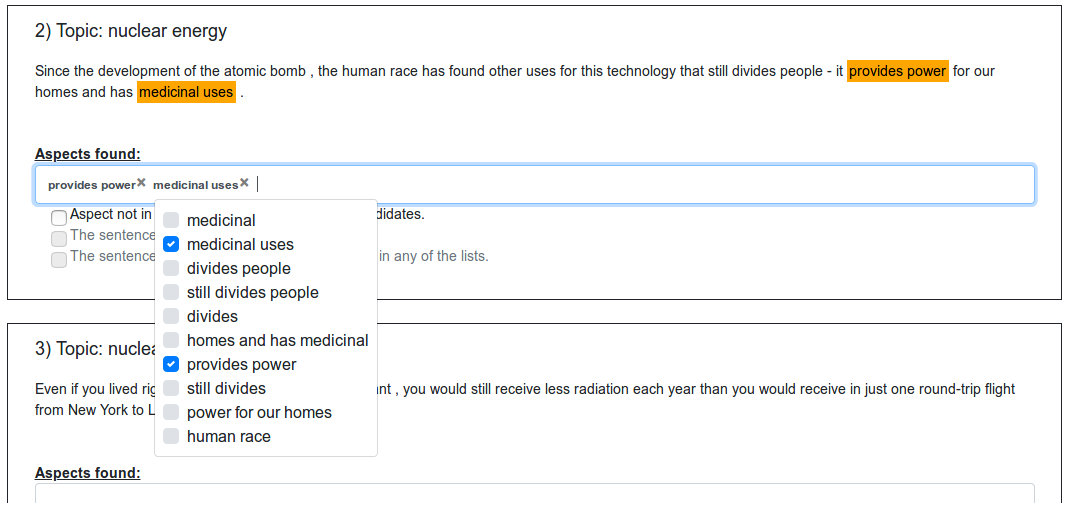}
    \caption{Example sentence of a HIT with the list of ranked aspect candidates.}
    \label{fig:aspect_amt_ex2}
\end{figure*}

Figure \ref{fig:aspect_amt_guidlines} shows the annotation guidelines for the final Amazon Mechanical Turk study. Figure \ref{fig:aspect_amt_ex1} shows one example of a HIT with two aspects selected. Selected aspects are highlighted in the sentence. We did not allow to choose overlapping aspects. If the aspect was not found in the first list provided by the learned ranker, the crowdworkers could choose from the remaining 1-4-grams of the sentence (aspect candidates starting or ending with stopwords, as well as candidates with punctuation and numbers, were removed from the list). Additional checkboxes were added to choose from if the sentence contained no aspect or the aspect was not explicitly mentioned. Figure \ref{fig:aspect_amt_ex2} shows a ranked list of aspect candidates for an example.

\subsection{Search Query and Topic Relevance Synonyms}\label{sec:synonyms}
In Table \ref{tbl:topic_rel_synonyms}, we show the synonyms used for preprocessing prior to the argument and stance classification with the ArgumenText API. We filtered out all sentences that did not contain tokens from the topic they belong to or any synonyms for this topic. Table \ref{tbl:search_query_synonyms} lists the ElasticSearch queries we used to retrieve the initial training documents from \cc and \reddit. Combinations of topics and data sources that are not listed in the table required no expansion of the query to gather enough documents for training.

\subsection{Reference Data Statistics}\label{sec:ref_data_stats}
Table \ref{tbl:reference_data_stats} shows the sources and number of arguments for all topics of the reference dataset. The dataset is used to compare the argument generation models to a retrieval approach.

\subsection{Examples of Generated Arguments}\label{sec:ex_generated_quality}
For all eight topics, we show the generated argument with the highest and lowest argument quality score in tables \ref{tbl_ex_quality_long_reddit} (\reddit model) and \ref{tbl_ex_quality_long_cc} (\cc model). Text in bold shows the given control code, text afterwards represents the generated argument. Numbers in brackets after the text shows the quality score as predicted by the argument quality model.

\section{Supplemental Material}
\label{sec:supplemental}
\subsection{Argument Aspect Detection Dataset}
The argument aspect detection dataset contains a total of 5,032 samples in JSONL-format, i.e. each dataset sample is in a separate line and can be parsed as JSON. A sample contains the keys:
\begin{itemize}
    \item \textbf{hash}: Unique identifier.
    \item \textbf{aspect\_pos}: List of string tuples ``(begin,length)'', marking the character position and length of each aspect within the argument.
    \item \textbf{aspect\_pos\_string}: The aspects as a list of strings.
    \item \textbf{stance}: Stance of the argument towards the topic, taken from the original dataset \cite{Stab2018b}.
    \item \textbf{topic}: The topic of the argument.
    \item \textbf{sentence}: The argument.
\end{itemize}
For reproducibility, we create a fixed cross-topic split with the data of two topics as test set (\textit{gun control}, \textit{school uniforms}), the data of one topic as development set (\textit{death penalty}), and the data of the remaining five topics as train set. We also create a fixed in-topic split.

\begin{table*}
\centering 
\resizebox{1.0\textwidth}{!}{
\begin{tabular}{ll}
\Xhline{2\arrayrulewidth}
\textbf{Topic} & \textbf{Synonyms} \\\hline
    School uniforms & uniform, college, outfit, dress, suit, jacket, cloth\\
    Nuclear energy & fission, fusion, atomic energy, nuclear power, atomic power, radioactive, radioactivity\\
    Marijuana legalization & cannabis, legalization of marijuana, legal, illegal, law, weed, dope\\
    Cloning & clone, cloned, duplicate, copy, reproduct, asexual \\
    Death penalty & capital punishment, execution, electric chair, punishment, punish\\
    Minimum wage & living wage, base pay, average wage, low income\\
    Abortion & abort, termination, misbirth, birth control\\
    Gun control & second amendment, ownership, arms reduction, arms limitation\\
\Xhline{2\arrayrulewidth}
\end{tabular}   
}
\caption{Topic synonyms to pre-filter sentences prior to argument and stance classification.}
\label{tbl:topic_rel_synonyms}
\end{table*}

\begin{table*}
\def\arraystretch{1.0}

\begin{tabularx}{\linewidth}{lX}
\Xhline{2\arrayrulewidth}
\textbf{Topic} & \textbf{Search query} \\\hline
\makecell[l]{Marijuana legalization\\(\cc and \reddit)} & ((marijuana legalization) OR (legalization of marijuana) OR (legalization of cannabis)) OR (((marijuana) OR (dope) OR (cannabis) OR (weed)) AND ((law) OR (legal) OR (legalization)))\\

\makecell[l]{School uniforms\\(\cc and \reddit)} & (school uniform) OR (college uniform) OR (school outfit) OR ((school) AND (uniform)) OR ((school) AND (outfit)) OR ((school) AND (jacket)) OR ((school) AND (cloth)) OR ((school) AND (dress)) OR ((college) AND (dress))\\

\makecell[l]{Cloning\\(\reddit)} & (clone) OR (cloning) OR (clones) OR (cloned) OR (clones) OR (genetically identical) OR (asexual reproduction)\\

\makecell[l]{Death penalty\\(\reddit)} & (capital punishment) OR ((execution) AND (death)) OR ((punishment) AND (death)) OR ((punishment) AND (execution)) OR (electric chair) OR (death penalty)\\

\makecell[l]{Gun control\\(\reddit)} & (gun control) OR (second amendment) or (2nd amendment) OR ((gun) AND (ownership))  OR (gun ownership) OR (arms reduction) OR (arms limitation) OR (gun politics) OR ((gun) AND (lobby))\\

\makecell[l]{Nuclear energy\\(\cc and \reddit)} & nuclear AND (energy OR fission OR power OR plant)\\

\Xhline{2\arrayrulewidth}
\end{tabularx}   
\caption{Synonyms added to the topic query to gather initial training documents from ElasticSearch. For combinations of topics and data sources (\reddit and \cc) that are not listed, we only used the topic as search query.}
\label{tbl:search_query_synonyms}
\end{table*}

\begin{table*}
\centering 
\def\arraystretch{1.0}
\begin{tabular}{lccc}
\Xhline{2\arrayrulewidth}
\textbf{Topic} & \multicolumn{2}{c}{\textbf{Sentence-Level Args}} & \textbf{Source} \\
& \textbf{pro} & \textbf{con} & \\\hline
Abortion & 57 & 47 & procon.org\\
Gun control & 49 & 50 & procon.org\\
School uniforms & 60 & 55 & procon.org\\
Minimum wage & 59 & 64 & procon.org\\
Marijuana legalization & 63 & 71 & procon.org\\
Death penalty & 22 & 27 & idebate.org\\
Cloning & 23 & 24 & idebate.org\\
Nuclear energy & 16 & 17 & idebate.org\\\hline
\textbf{Total} & \textbf{349} & \textbf{355} & /\\
\Xhline{2\arrayrulewidth}
\end{tabular}   
\caption{Sources and number of arguments for each topic from the reference dataset.}
\label{tbl:reference_data_stats}
\end{table*}

\begin{table*}[!hbt]
\small
\begin{tabularx}{\linewidth}{|X|}
\hline
\textbf{abortion CON violence :} Abortion is not only morally wrong but also physically dangerous for both mother and baby ; it increases breast cancer risk by 50 \% and causes mental illness including schizophrenia and psychosis ; it leads to infertility and premature birth in subsequent pregnancies ; it contributes significantly to maternal mortality due to hemorrhage during childbirth ; complications arising from unsafe abortion account for 13 percent of all maternal deaths worldwide ; and post-abortion syndrome affects up to one quarter of pregnant women suffering from this condition . (0.99)\\
\textbf{abortion CON lives .} ..I 'm sorry , but your logic does n't pass the laugh test . (0.49)\\\hline

\textbf{cloning CON unrespectable .} Cloning humans for reproductive purposes is unethical and unacceptable , but creating cloned embryos solely for research - which involves destroying them anyway - is downright criminal . (0.97)\\
\textbf{cloning CON disfavored .} , cliques ) to them . (0.36)\\\hline

\textbf{nuclear energy PRO safe .} In addition , we must continue developing safer technologies like small modular reactors which will help us meet our nation 's need for reliable , emission-free sources of low-emission energy while also creating jobs and providing solutions to some of the world s most pressing problems : climate change , food security and sustainable development . (0.96)\\
\textbf{nuclear energy CON leak .} `` We are concerned about the possibility of further releases of radioactivity due to possible melting or cracking of fuel rods at the No . (0.47)\\\hline

\textbf{death penalty CON inhuman .} Amnesty International opposes the death penalty in all cases as the ultimate form of cruel , inhuman or degrading punishment and a violation of fundamental rights - the right to life and the prohibition of torture . (1.00)\\
\textbf{death penalty CON god .} And yet , while the Church exhorts civil authorities . (0.41)\\\hline

\textbf{school uniforms CON resources .} The lack of a uniform system for the collection and use of data , as well as insufficient funding to collect it , are major barriers that limit access to information on student achievement in schools with high concentrations of students from low-income families . (0.92)\\
\textbf{school uniforms PRO fun :} The kids are having a lot of fun . (0.38)\\\hline

\textbf{gun control PRO homicides .} In addition to being an effective crime deterrent and reducing suicides , research has shown that defensive firearm use prevents more than 80 percent of all gun-related violent crimes including homicides , robberies , assaults , rapes , and aggravated assaults [ 1 ] . (0.96)\\
\textbf{gun control PRO prevent ;} You liberals need to wake up and smell the coffee . (0.43)\\\hline

\textbf{marijuana legalization PRO safer :} Legalizing cannabis will help reduce crime rates ( especially violent crimes ) and make society safer overall . (0.96)\\
\textbf{marijuana legalization PRO benefits .} Decrease amount of police officers needed 6 . (0.37)\\\hline

\textbf{minimum wage PRO poor :} Raising the minimum wage will not only benefit those working full time but also reduce government expenditures on social services such as food stamps and Medicaid which disproportionately affect the poor . (0.97)\\
\textbf{minimum wage CON cost :} If you raise the price of a Big Mac to \$ 10 and then pay an extra dime or two per burger so that it 's still only \$ 9 ... well , maybe your business is n't worth saving at all because there just wo n't be anyone willing to eat out after they 've had their fill ( unless perhaps they 're on welfare ) and now have no job and ca n't afford health care either - but I doubt very seriously that most people making those kinds of wages/salaries would suddenly decide not to work if forced to shoulder the added cost . (0.44)\\\hline
                                  
\end{tabularx}
\caption{Generated arguments with \cc as data source. Text in bold shows the given control code, text afterwards represents the generated argument. Numbers in brackets after the text shows the quality score as predicted by the argument quality model.}\label{tbl_ex_quality_long_reddit}
\end{table*}

\begin{table*}[!hbt]
\small
\begin{tabularx}{\linewidth}{|X|}
\hline
\textbf{abortion PRO reduce abortions :} It 's been shown time after time that comprehensive sexual health services ( including free or cheap long term hormonal birth controls ) greatly reduces abortions by preventing unplanned pregnancies from happening in the first place . (0.99)\\
\textbf{abortion PRO crime .} \_r=0 \& amp ; pagewanted=print \& amp ; oref=slogin ) . (0.40)\\\hline

\textbf{cloning PRO reproduction .} The only way to increase the number of clones is through sexual reproduction , which increases genetic diversity and therefore reduces extinction rates . (0.85)\\
\textbf{cloning PRO awesome .} But yeah , the clone skins look fucking awesome . (0.36)\\\hline

\textbf{nuclear energy PRO safe .} Nuclear is the only viable option for a large scale , reliable and safe form of energy production that can replace fossil fuels as our main energy source . (0.97)\\
\textbf{nuclear energy CON leak .} Biofuel does n't need batteries 6 . (0.41)\\\hline

\textbf{death penalty PRO save .} The only way we can possibly make sure no innocents are executed is by abolishing the death penalty altogether - there 's just too much chance that at least one innocent person will die before their execution date was up and they were able to prove their innocence with DNA evidence and/or other exonerating circumstances . (0.95)\\
\textbf{death penalty PRO innocent person .} Innocent people do n't deserve to live 2 . (0.43)\\\hline

\textbf{school uniforms PRO fit .} Dress codes exist to prevent distractions from other students while trying to teach kids appropriate attire which helps them learn proper social skills and fitting into society . (0.83)\\
\textbf{school uniforms PRO nice :} It looks really nice on my college application . (0.37)\\\hline

\textbf{gun control PRO prevent .} Guns also help prevent tyranny by removing checks against government overreach into areas where the populace has little power . (0.95)\\
\textbf{gun control CON problem ;} the guns are n't the real problems . (0.32)\\\hline

\textbf{marijuana legalization CON bad :} Alcohol is also very addictive and has been shown time after time to have negative effects on health yet it remains completely legal while cannabis gets demonized by law enforcement and politicians alike despite being less harmful than many prescription medications in every way imaginable . (0.93)\\
\textbf{marijuana legalization PRO buy .} Get busted by police 5 . (0.36)\\\hline

\textbf{minimum wage PRO poverty :} Raising the minimum wage helps alleviate poverty as well as increase demand for goods and services from consumers . (0.93)\\
\textbf{minimum wage CON pay :} They ca n't pay below minimum wage either . (0.41)\\\hline
                                  
\end{tabularx}
\caption{Generated arguments with \reddit as data source. Text in bold shows the given control code, text afterwards represents the generated argument. Numbers in brackets after the text shows the quality score as predicted by the argument quality model.}\label{tbl_ex_quality_long_cc}
\end{table*}
\end{document}